%% file: icme2022.tex
\documentclass{article}
\usepackage{amsmath,epsfig}
\usepackage{graphicx,amssymb,multirow,hyphenat,float,subcaption,booktabs}
\usepackage[preprint]{spconfa4}

\copyrightnotice{Copyright notice – please select from the list provided at Camera-Ready Instructions}

\let\OLDthebibliography\thebibliography
\renewcommand\thebibliography[1]{
  \OLDthebibliography{#1}
  \setlength{\parskip}{0pt}
  \setlength{\itemsep}{0pt plus 0.3ex}
}

\begin{document}\sloppy

% Example definitions.
% --------------------
\def\x{{\mathbf x}}
\def\L{{\cal L}}

% Title.
% ------
\title{Online Deep Metric Learning via Mutual Distillation}
%
% Address.
% ---------------
\name{Gao-Dong Liu$^{\ast}$, Wan-Lei Zhao$^{\ast}$, Jie Zhao$^{\dagger}$}
\address{$^{\ast}$Xiamen University; $^{\dagger}$Boden.ai.}

\maketitle

\begin{abstract}
  Deep metric learning aims to transform input data into an embedding space, where similar samples are close while dissimilar samples are far apart from each other. In practice, samples of new categories arrive incrementally, which requires the periodical augmentation of the learned model. The fine-tuning on the new categories usually leads to poor performance on the old, which is known as ``catastrophic forgetting''. Existing solutions either retrain the model from scratch or require the replay of old samples during the training. In this paper, a complete online deep metric learning framework is proposed based on mutual distillation for both one-task and multi-task scenarios. Different from the teacher-student framework, the proposed approach treats the old and new learning tasks with equal importance. No preference over the old or new knowledge is caused. In addition, a novel virtual feature estimation approach is proposed to recover the features assumed to be extracted by the old models. It allows the distillation between the new and the old models without the replay of old training samples or the holding of old models during the training. A comprehensive study shows the superior performance of our approach with the support of different backbones.
  
  \end{abstract}
  \input{intro.tex}
  \input{relate.tex}
  \input{method.tex}
  \input{exp.tex}
  \input{conclude.tex}

% References should be produced using the bibtex program from suitable
% BiBTeX files (here: strings, refs, manuals). The IEEEbib.bst bibliography
% style file from IEEE produces unsorted bibliography list.
% -------------------------------------------------------------------------
\bibliographystyle{IEEEbib}
\bibliography{metric}

\end{document}

%% file: intro.tex
\section{Introduction}
Owing to the seminal learning framework from~\cite{chopra2005learning}, deep metric learning has been successfully applied in various tasks such as face recognition~\cite{chopra2005learning,schroff2015facenet}, person re-identification~\cite{yi2014deep}, and fine-grained image search~\cite{oh2016deep,wang2019multi}, etc. The research focus in recent years has been on mining hard training samples~\cite{schroff2015facenet,oh2016deep,wu2017sampling}. In most of these works, the visual categories to be trained in the training set are fixed. No mechanism is designed to allow new categories to join in the training incrementally. Only a few research  works~\cite{chen2021feature,chen2020exploration,tian2020complementary,wu2019deep} shed light on this learning issue, which is known as ``online deep metric learning'', or ``incremental deep metric learning''. In this scenario, the trained model has to be updated to adapt to new categories on the one hand. On the other hand, it is required to maintain the performance on old categories as much as possible. Due to the well-known ``catastrophic forgetting''~\cite{catastrophic}, these two competing requirements are hardly balanced.

There are several possible practices for the online deep metric learning. An intuitive solution is to retrain a new model based on both the old and new categories, which is also known as joint training. The major disadvantage of this type of approaches is that it requires large working memory to store and replay the past training samples~\cite{shin2017continual,rebuffi2017icarl}. Such storage and replay may not be viable in practice. For instance, the samples of old categories are no longer available for streaming data. Another way is to fine-tune the trained model with the samples of new categories only. The embedding space constructed based on old categories has been transformed to adapt new categories, which leads to the considerable degradation on old tasks. This is essentially the cause of ``catastrophic forgetting''. 

In the literature, most of the online learning is addressed under the context of visual class categorization task. In~\cite{shin2017continual}, the new model is trained based on the samples from both new and old categories that are produced by a generator. It requires the replay of samples from old categories. Essentially, it is still a joint training approach based on the generative model. Recent works~\cite{kirkpatrick2017overcoming,li2017learning} address this issue under a more stringent condition where old data is not available during the training of new tasks. Research works in~\cite{chen2021feature,chen2020exploration} explore the online deep metric learning under the same constraint. Although different frameworks have been proposed, they all intend to maintain the distribution of old categories the same as before in the augmented embedding space. As no old data is used, different ways have been introduced to recover the features produced by the old models. The recovered old features will facilitate the training of a new model to keep balanced performance on both old and new tasks.

In all of the aforementioned approaches~\cite{chen2021feature,chen2020exploration, kirkpatrick2017overcoming,li2017learning}, in order to maintain the knowledge learned from the old tasks, the performance in the new tasks has been sacrificed. In this paper, a novel online deep metric learning approach is proposed based on mutual learning~\cite{zhang2018deep}, where {\textit{three}} models are involved to strike a balance between stability and plasticity. \textit{Two} student models learn collaboratively throughout the training, leading to the better generalization to the new task~\cite{zhang2018deep}. \textit{One} teacher model transfers previous knowledge to student models to preserve the performance on the old task. Moreover, we extend this one-stage online learning framework to the scenario of multiple stages, where new categories are allowed to join in multiple batches. In order to enable the mutual learning of knowledge from earlier stages, virtual features which are assumed to be produced from previous models are estimated. This allows the acquired knowledge to be transferred from previous stages to the current stage without the replay of old training samples or loading of old models.

%% file: relate.tex
\section{Related Work}
\label{sec:relate}
Online deep metric learning has been addressed under different assumptions in the literature. In the case that the replay of old training samples is allowed~\cite{tian2020complementary,wu2019deep}, it is essentially a variant of joint training. The drawback is that the old training samples are not always available due to the issue of privacy concerns and the high maintenance costs. Under the more widely recognized assumption, the old training samples are not allowed to join in the training of new tasks. Recent approaches~\cite{chen2021feature,chen2020exploration} are all proposed under this assumption. Although different in the design of loss functions, they all regularize the new model with the old ones to preserve the inherent feature distribution that is learned from the old data.

Usually, knowledge distillation~\cite{hinton2015distilling} is adopted to distill the feature information of the old categories from the old model to the new model~\cite{chen2021feature,chen2020exploration}. Intuitively, the current model inherits all the trained categories from the previous models. The last model preserves the feature distribution of all the previously trained categories. It is possible to correlate the feature distribution of all the previous categories based on the latest model. However, the errors caused by the incremental learning could be aggregated. The discriminativeness on the old tasks is therefore eroded gradually. As a result, recovering the feature space of the previous stages is still necessary. In~\cite{chen2020exploration}, a Maximum Mean Discrepancy (MMD) based regularization loss is introduced to minimize the discrepancy between features of newly added categories from the original and adaptive networks. In~\cite{chen2021feature}, features from models of previous stages are estimated. The old features are recovered according to the variation of mAP before and after the training on the new task. The disadvantages of this approach lie in two aspects. Firstly, not all the variations in the feature space can be reflected by the accuracy variation in the image retrieval task. Moreover, the variation of feature space cannot simply be modeled as a linear transformation.

Mutual learning~\cite{zhang2018deep}, as an extension of knowledge distillation, is an ensemble training strategy to improve generalization by transferring individual knowledge to each other. In such kind of framework, it is not required to have a fully-trained teacher model. Instead, two peer student models are trained and learn from each other via a mutual loss. As the framework shows no preference over any individual model, neither old nor new knowledge will dominate over each other.

In this paper, the online deep metric learning is addressed under the assumption that the replay of old training samples is not allowed. Different from existing approaches, mutual learning instead of a teacher-student framework is adopted for knowledge transfer from one stage to the next. We will show that it is more suitable for online deep metric learning. Additionally, a novel feature estimation strategy is proposed. Based on the models of previous stages, the drifting of feature space during the training of multiple stages can be captured. This in turn allows us to recover the feature distribution in old models more precisely than that of FECD~\cite{chen2021feature}.

%% file: method.tex
\section{Method}
\label{sec:method}
\subsection{Problem Formulation}
In the real world, the scale of learning issues increases incrementally on many occasions. For example, for a shopping website, more and more categories of products are put on sale periodically. It is required the trained model to be updated periodically. Given one stage of the incremental training is defined as one training task $\tau$, the online deep metric learning is composed of a sequential set of training tasks $T=\{\tau_1,\tau_2,\cdots,\tau_i\,\cdots\}$. In one training task $\tau_i$, it consists of a training set with $n$ new categories, namely $\tau_i=\{(X^c_i, y^c_i)|c=1,2,...,n\}$, where $X^c_i$ are the set of training samples $x$. All the samples in $X^c_i$ share the same class label $y^c_i$. Without loss of generality, we assume there is no intersection between any two training tasks, $\tau_i \cap \tau_j = \emptyset$.\footnote{Otherwise, the training on the overlapped categories is a fine-tuning of the trained model.} Correspondingly, we expect a series of models are learned with the given training sets, namely $M=\{F_1(X^c_1, w_1),F_2(X^c_2, w_2),\cdots,F_i(X^c_i, w_i),\cdots\}$, where $w_i$ are the trained weights of a model $F_i$. For each trained model $F_i$, it is essentially a mapping function, through which a given image $x$ is mapped to a fixed-length feature vector $f_x^i$.

On the condition that the old training sample replay is not allowed, model $F_i$ could be trained in two different ways. In the first way, all the previous models $F_1, F_2, \cdots, F_{i-1}$ along with the training set $\tau_i=\{(X^c_i, y^c_i)\}$ are available. Alternatively, only model $F_{i-1}$ and $\tau_i=\{(X^c_i, y^c_i)\}$ are available for training task $\tau_i$. In our solution, the online deep metric learning is addressed in the first way. In the following, we are going to first present a solution for the online deep metric learning that only involves two training tasks. We assume the initial model $F_{o}(X^c_o, w_o)$ is well trained on task $\tau_o$ with \textit{n} old categories ${(X^{c}_o, y^{c}_o)}$. The weights $w_o$ already converges on task $\tau_o$. Now new task $\tau_p$ with \textit{m} new categories ${(X^{c}_p, y^{c}_p)}$ are joined in. This is called ``one-task online learning''. Upon the basis of one-task online learning, the solution to the multi-task online learning is presented in the section followed.

\subsection{One-task Online Learning}
\begin{figure}[tb]
\begin{center}
   \includegraphics[width=0.81\linewidth]{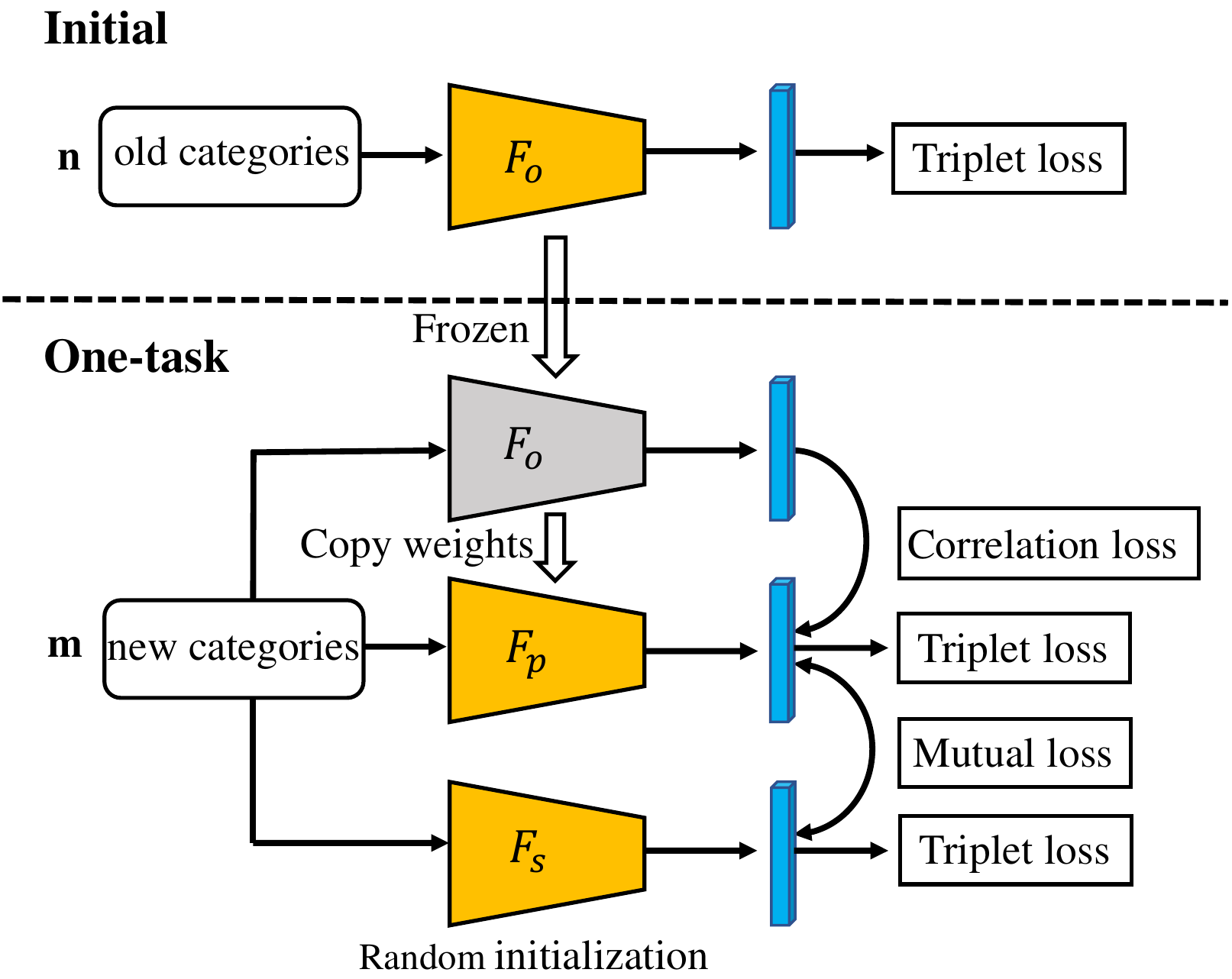}
   \caption{One-task online learning overview. Initial: A model $F_{o}$ is trained on $n$ old categories ${(X^{c}_o, y^{c}_o)}$. One-task: $F_{o}$ is frozen as the teacher model and duplicated as the initialization of the student model $F_{p}$. A randomly initialized model $F_{s}$ as the supporting student model is employed to solve the new task with $F_p$. Only samples of $m$ new categories ${(X^{c}_p, y^{c}_p)}$ are available during the training.}
\label{fig:onetask}
\end{center}
\end{figure}

A two-student mutual learning framework~\cite{zhang2018deep} is adopted in our one-task online learning. The framework is shown in Fig.~\ref{fig:onetask}. Basically, there are three branches in the framework. The first branch copies the model trained at the ``Initial'' stage and its weights are frozen during the training. Given a training sample $x$, it produces a feature, namely  $f^o_x=F_o(x)$, which is used as a reference during the training of $F_p$.

The second branch initially copies weights from the first branch, namely $w_p \leftarrow w_o$. It is designed to train a model $F_p$ that maintains the discriminativeness on the old categories and adapts to the new categories in $\tau_p$. The third branch is a supporting model that is initialized with random weights. On the one hand, $F_s$ learns the new categories from scratch. The triplet loss is adopted in $F_p$ and $F_s$ to separately learn the new categories in task $\tau_p$. On the other hand, $F_s$ and $F_p$ also learn from each other, which learned knowledge is shared by the introduction of mutual loss on both branches. All these three branches share the same network structure. $F_p$ and $F_s$ in combination are called student models~\cite{zhang2018deep}. $F_o$ is as the teacher model. The objective for one-task online learning is

\begin{equation}
\begin{aligned}
L(X^{c}_p;w_{o};w_{p};w_{s}) &= \lambda_1L_{triplet}(X^{c}_p;w_{p})  \\
&+ \lambda_1L_{triplet}(X^{c}_p;w_{s})  \\
&+ \lambda_2L_{corr}(X^{c}_p;w_{o};w_{p}) \\
&+ \lambda_3L_{mutual}(X^{c}_p;w_{s};w_{p}),
\end{aligned}
\label{eqn:lbranch1}
\end{equation}
where $\lambda_1$, $\lambda_2$, and $\lambda_3$ are the hyperparameters used for weighting of different loss functions. In Eqn.~\ref{eqn:lbranch1}, $\lambda_1$ is set to \textit{1} and $\lambda_2$ is set to \textit{10}~\cite{chen2021feature}. $\lambda_3$ is empirically set to \textit{8}, of which leads to the best performance according to our observation. $L_{triplet}$ is the triplet loss using hard sampling strategy. $L_{corr} = \frac{1}{N}\sum KL(\sigma(G_o), \sigma(G_p))$ is the correlation distillation loss between $F_o$ and $F_p$~\cite{chen2021feature}. It applies Kullback\hyp{}Leibler divergence between two Gram matrices $G_o, G_p$ which are calculated with features extracted by $F_o, F_p$ and normalized with Softmax function $\sigma(\cdot)$. $F_{s}$ also generates a Gram matrix $G_{s}$ to regularizes the updating of $F_{p}$. The mutual distillation loss is defined as follows
\begin{equation}
\begin{aligned}
        L_{mutual-p} &= \frac{1}{N}\sum KL(\sigma(G_p), \sigma(G_s))\\
        L_{mutual-s} &= \frac{1}{N}\sum KL(\sigma(G_s), \sigma(G_p))\\
        L_{mutual} &= \frac{1}{2}(L_{mutual-p} + L_{mutual-s}).
\end{aligned}
\label{eqn:mutual}
\end{equation}
According to Eqn.~\ref{eqn:lbranch1}, $F_p$ is required to mimic $F_o$ on the one hand. On the other hand, $F_p$ also learns from $F_s$, which is better trained on the new task $\tau_p$. Since $F_s$ also learns from $F_p$, $F_p$ and $F_s$ converge to similar models as the training continues. Compared to the popular teacher-student distillation framework, mutual distillation helps us to find a wider/flatter robust minimum that generalizes better to the new task. In the multi-task scenario where tasks are added incrementally, performing online learning from flat minima will effectively mitigate forgetting of previous tasks~\cite{shi2021overcoming}. In the following, the online deep metric learning is addressed in the multi-task context based on our one-task online learning solution.

%\begin{figure}[tb]
%\begin{subfigure}{0.5\linewidth}
%  \centering
%  \includegraphics[width=\linewidth]{figs/feature-estimation-1.pdf}
%  \caption{}
%  \label{subfig:prototype}
%\end{subfigure}\hfill
%\begin{subfigure}{0.5\linewidth}
%  \centering
%  \includegraphics[width=\linewidth]{figs/feature-estimation-2.pdf}
%  \caption{}
%  \label{subfig:feature}
%\end{subfigure}
%\caption{Illustration of feature estimation for prior tasks. (a). Feature drifts are utilized to approximate the prototype drifts of task $b$ and update learned prototypes. (b). Estimating virtual features of task $b$ with features extracted by model $f_{t-1}$, $y_i \in C^{t}$.}
%\label{fig:estimation}
%\end{figure}

\begin{table*}[t]
\centering
\scriptsize{
\begin{tabular}{lcccccc} 
\toprule
\multirow{2}{*}{Datasets} & \multicolumn{3}{c}{\begin{tabular}[c]{@{}c@{}}Training set\\(\#Image/\#Class)\end{tabular}} & \multicolumn{3}{c}{\begin{tabular}[c]{@{}c@{}}Testing set\\(\#Image/\#Class)\end{tabular}}  \\
                          & Old task & New task & All & Old task & New task & All \\ 
\hline
CUB-200                   & 3,504/100    & 3,544/100    & 7,048/200                                                        & 2,360/100    & 2,380/100    & 4,740/200                                                        \\
Cars196                   & 4,796/98     & 4,842/98     & 9,638/196                                                        & 3,258/98     & 3,289/98     & 6,547/196                                                        \\
DeepFashion2              & 54,898/10,595 & 55,532/10,595 & 110,430/21,190                                                    & 43,615/10,595 & 44,095/10,595 & 87,710/21,190                                                     \\
\bottomrule
\end{tabular}}
\caption{Statistics of three datasets.}
\label{table:datasets}
\end{table*}

\subsection{Multi-task Online Learning}
In practice, new categories arrive in multiple stages. The model has to be trained periodically. At each stage, the model is trained on an aggregated new task. Namely, given there are $i-1~(i-1 \geq 2)$ tasks have been trained, we are now going to integrate a new group of categories $\tau_i=\{(X^c_i, y^c_i)|c=1,2,...,n\}$ into the trained model $F_{i-1}$ and work out a new model $F_i$. Although all the trained categories are kept in model $F_{i-1}$, the feature correlation for early tasks is distorted due to the aggregated concept drifting, which is more serious for the earlier tasks. It is, therefore, insufficient to correlate model $F_i$ with $F_{i-1}$ for the earlier tasks. Hereby, we assume all the models trained on the previous tasks are available. These models will be used to guide the more precise correlation of $F_i$ with them.

Given a category $k$ in $\tau_b~(b < i-1)$, prototype $\mu^b_{b,k}$ is defined as the centroid of $k$th category in task $\tau_b$, which is calculated by the mean of features in this category extracted by $F_b$. When our training reaches Stage-$i$, the distribution of category $k$ has drifted already. On the one hand, the image sample $x \in \tau_i$ should be fed into $F_b$ to supervise the correlation of $F_i$ to $F_b$. However, it is computationally expensive to load all the previous models to support the feature correlation and there is only one loaded previous model $F_{i-1}$. On the other hand, the feature from $F_b$ cannot be directly estimated based on $F_{i-1}$ since the drift of feature between $F_b$ and $F_{i-1}$ is unknown. Hereby, a way to estimate this drift is proposed. 

Given a training sample $x~(x \in \tau_{i-1})$, features $f^b_x$ and $f^{i-1}_x$ are extracted from $F_{b}~(b=1{\cdots}i-2)$ and $F_{i-1}$ respectively\footnote{This operation is undertaken before we train $F_i$. It is, therefore, offline.}. The drift of this individual feature is ${{\Delta}f}^{b\xrightarrow{}i-1}_x= f^{i-1}_x - f^b_x$. Following with~\cite{yu2020semantic}, the drift of the prototype $\mu^b_{b,k}$ at Stage-$i-1$ can be estimated as
\begin{equation}
{{\Delta}{\mu}}^{b\xrightarrow{}i-1}_{b,k} = \frac{\sum_x{S(f^b_x,{\mu}^b_{b,k}) {{\Delta}f}^{b\xrightarrow{}i-1}_x}}{\sum_x {S(f^b_x,{\mu}^b_{b,k})}}, x \in \tau_{i-1},
\label{eqn:prototype-drift}
\end{equation}
where $S(f^b_x,\mu^b_{b,k})$ is the \textit{Cosine} similarity between the feature $f^b_x$ and the learned prototype $\mu^b_{b,k}$ at Stage-$b$. As we can see from Eqn.~\ref{eqn:prototype-drift}, the overall drift between two prototypes ($\mu^b_{b,k}, \mu^{i-1}_{b,k}$) is estimated by the weighted drift of each individual feature. Therefore, the prototype for category $k~ (k \in \tau_b)$ at Stage-$i-1$ is updated as ${\mu}^{i-1}_{b,k} = \mu^b_{b,k} + {{\Delta}{\mu}}^{b\xrightarrow{}i-1}_{b,k}$.

During the training of Stage-$i$, only the previous model $F_{i-1}$ is loaded as the teacher model. For a previous task $\tau_b$,  all of the prototype drifts ${{\Delta}{\mu}}^{b\xrightarrow{}i-1}_{b,k=1 \cdots n}$ are calculated and kept for updating the prototypes ${{\mu}^b_{b,k=1 \cdots n}\xrightarrow{}{\mu}^{i-1}_{b,k=1 \cdots n}}$. Given a training sample $x \in \tau_i$, its feature $f^b_x$ on $F_b$ is not extracted directly as $F_b$ is not loaded. Instead, it is estimated based on the teacher feature $f^{i-1}_x$. Namely, given $f^{i-1}_x$, ${\mu}^{i-1}_{b,k=1 \cdots n}$ and ${{\Delta}{\mu}}^{b\xrightarrow{}i-1}_{b,k=1 \cdots n}$, we estimate how much $f^{i-1}_x$ deviates from $f^b_x$
\begin{equation}
{{\Delta}f}^{i-1\xrightarrow{}b}_x = - \frac{\sum_{k=1}^{n} {S(f^{i-1}_x, {\mu}^{i-1}_{b,k}) {{\Delta}{\mu}}^{b\xrightarrow{}i-1}_{b,k}}}{\sum_{k=1}^{n} {S(f^{i-1}_x,{\mu}^{i-1}_{b,k})}}, k \in \tau_b.
\label{eqn:feature-drift}
\end{equation}
Similar as Eqn.~\ref{eqn:prototype-drift}, Eqn.~\ref{eqn:feature-drift} calculates a weighted drift for the feature $f^{i-1}_x$ with respect to all the drifted prototypes and their drifts at Stage-$i-1$. Consequently, virtual feature $\hat{f}^b_x$ is estimated as $\hat{f}^b_x = f^{i-1}_x + {{\Delta}f}^{i-1\xrightarrow{}b}_x$. These estimated virtual features at Stage-$b$ regularize the updates of the current model $F_i$ by constructing a correlation loss
\begin{equation}
   L_{corr}^{b\xrightarrow{}i} = \frac{1}{N}\sum KL(\sigma(G_{b}),\sigma(G_{i})).
   \label{eqn:correlation-example}
\end{equation}

\begin{figure}[tb]
   \centering
   \includegraphics[width=0.7\linewidth]{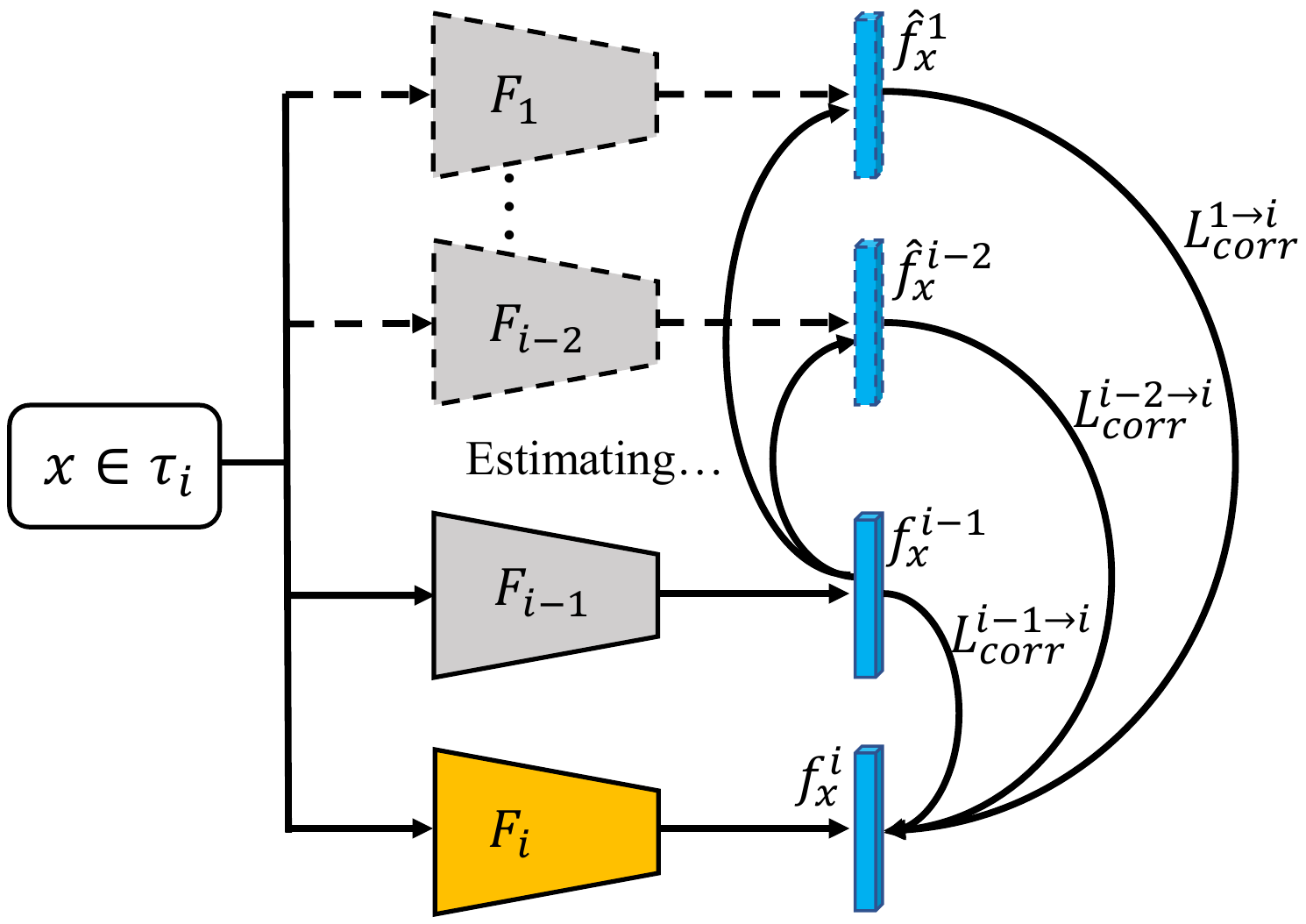}
   \caption{Correlation regularization in multi-task online learning. Features of prior tasks are estimated from the teacher feature instead of extracting them with prior models. The Gram matrices produced by virtual features are used as supervision for learning the current task.}
   \label{fig:multitask}
\end{figure}

As shown in Fig.~\ref{fig:multitask}, we further estimate virtual features for tasks $\tau_1, \tau_2, \cdots, \tau_{i-2}$. With these virtual features, more Gram matrices are built, which provide more auxiliary knowledge rather than the only distilled knowledge from $F_{i-1}$ for the current task $\tau_i$
\begin{equation}
L_{corr} = \sum_t {L_{corr}^{t\xrightarrow{}i}}, t=1, 2, \cdots, i-1.
\label{eqn:correlation-all}
\end{equation}
During the training, the 3rd term in Eqn.~\ref{eqn:lbranch1} is replaced with Eqn.~\ref{eqn:correlation-all}, which becomes the overall loss function for multi-task online deep metric learning.

%% file: exp.tex
\section{Experiment}
\label{sec:exp}
In this section, the effectiveness of our online deep metric learning approach is studied on three datasets, Caltech-UCSD Birds 200 (CUB-200)~\cite{WelinderEtal2010}, Cars196~\cite{KrauseStarkDengFei-Fei_3DRR2013}, and DeepFashion2~\cite{ge2019deepfashion2}. The major information of these three datasets is summarized in Tab.~\ref{table:datasets}. In the one-task learning case, the training set and the testing set of three datasets are evenly divided into two halves. The first half categories are treated as the old task, and the second half categories are treated as the new task. The performance of our approach is compared to LwF~\cite{li2017learning}, EWC~\cite{kirkpatrick2017overcoming}, FGIR~\cite{chen2020exploration}, and FECD~\cite{chen2021feature}. For all the approaches, BN\hyp{}Inception~\cite{ioffe2015batch} is adopted as the backbone, and triplet loss is used as the basic loss function. The results from FECD~\cite{chen2021feature} are treated as the comparison baseline. The result from joint training is supplied as a reference, which learns all the categories as one training task. The performance is evaluated by Recall@1 for the retrieval on each dataset.

We noticed the experimental flaws pointed out by~\cite{musgrave2020metric} in the current literature of deep metric learning. In our experiment design, no sophisticated image augmentation is adopted. The parameters in Batch-Norm are frozen. The same optimizer \textit{Adam optimizer} is used for all the approaches during the training.

% \begin{table}[tb]
% \centering
% \scriptsize{
% \begin{tabular}{cccc} 

% \end{tabular}}
% \caption{Recall@1 (\%) on one-task online learning. ``Initial'', ``Fine-tuning'' and ``Joint Training'' are as the references.}
% \label{table:onetask-results}
% \end{table}

\begin{table}[tb]
\centering
\scriptsize{
\begin{tabular}{ccccc} 
\toprule
Datasets                      & Approaches        & Old            & New            & All             \\ 
\hline
\multirow{8}{*}{CUB-200}      & Initial        & 79.79          & 55.55          & 62.76           \\
                              & Fine-tuning    & 68.81          & 79.92          & 69.01           \\
                              & Joint Training & 77.67          & 79.66          & 76.75           \\
                              & LwF            & 54.92          & 75.76          & -               \\
                              & EWC            & 62.03          & 73.32          & -               \\
                              & FGIR           & 74.41          & 73.11          & -               \\
                              & FECD           & 77.20          & 76.09          & 72.43           \\
                              & \textbf{Ours}  & \textbf{77.29} & \textbf{78.07} & \textbf{73.73}  \\ 
\hline
\multirow{5}{*}{Cars196}      & Initial        & 82.23          & 63.58          & -               \\
                              & Fine-tuning    & 67.77          & 90.00          & -               \\
                              & Joint Training & 82.26          & 92.28          & 85.15           \\
                              & FECD           & 80.82          & 91.21          & 82.36           \\
                              & \textbf{Ours}  & \textbf{81.06} & \textbf{92.40} & \textbf{83.49}  \\ 
\hline
\multirow{5}{*}{DeepFashion2} & Initial        & 55.79          & 55.04          & -               \\
                              & Fine-tuning    & 54.81          & 54.97          & -               \\
                              & Joint Training & 56.02          & 56.21          & 49.66           \\
                              & FECD           & 56.26          & 56.49          & 50.18           \\
                              & \textbf{Ours}  & \textbf{56.33} & \textbf{56.65} & \textbf{50.35}  \\
\bottomrule
\end{tabular}}
\caption{Recall@1 (\%) on one-task online learning. ``Initial'', ``Fine-tuning'' and ``Joint Training'' are as the references.}
\label{table:onetask-results}
\end{table}

\subsection{Performance on One-task Online Learning}
The experimental results on three datasets are reported on Tab.~\ref{table:onetask-results}. The results for LwF, EWC, and FGIR are cited from~\cite{chen2020exploration} directly. The results from the ``Initial'' model and ``Fine-tuning'' model are reported. The ``Initial'' model is trained on the old task only. The ``Fine-tuning'' model is trained on the old task and then fine-tuned on the new task.

As shown on the table, the initial model and fine-tuned model perform poorly either on CUB-200 or Cars196. This does indicate the necessity of an online metric learning approach to fit in. Among all the online approaches, our approach outperforms the rest constantly on all three tasks (old, new, and all). The performance gap between offline and online approaches on DeepFashion2 is minor. This is mainly because the training task is in large-scale (\textit{21,190} categories). The discriminativeness of the trained model is already saturated after the first half categories have been trained. Nevertheless, our approach shows the constant improvement over existing approaches, as it shows the good trade-off between the old and new tasks.

\subsection{Performance on Multi-task Online Learning}
In order to simulate the online learning of multiple stages, the second half of the categories in three datasets are divided into four subsets evenly. So for CUB-200, Cars196, and DeepFashion2, there are \textit{25}, \textit{25}, and \textit{2,661} new categories respectively joining in the training as a new task at each stage. In order to be in line with~\cite{chen2021feature}, the performance of the model in multi-task online learning is reported in two ways. Firstly, the Recall@1 of the final model is reported on each individual stage. Moreover, the overall performance of the final model on all the trained categories is reported.

The performance from our approach and LwF~\cite{li2017learning}, EWC~\cite{kirkpatrick2017overcoming}, FGIR~\cite{chen2020exploration}, and FECD~\cite{chen2021feature} are reported in Tab.~\ref{table:multitask-results}. As shown from the table, FECD and our approach outperform the rest online approaches by a large margin. Compared to FECD, our approach shows constantly better performance in most of the stages. In particular, our approach outperforms FECD across all the tasks on Cars196 and DeepFashion2 datasets. The superior performance of our approach owes to both the adoption of mutual learning and novel virtual feature estimation for the old models. 

\begin{table}[tb]
\centering
\scriptsize {
\begin{tabular}{cccccccc} 
\toprule
Datasets                      & Tasks       & Joint & LwF   & EWC   & FGIR  & FECD           & \textbf{Ours}   \\ 
\hline
\multirow{6}{*}{CUB-200     } & 1-100       & 77.67 & 33.31 & 36.82 & 66.40 & 73.64          & \textbf{73.90}  \\
                              & 101-125     & 82.14 & 49.83 & 57.99 & 70.07 & \textbf{77.21}  & \textbf{77.21}   \\
                              & 126-150     & 77.83 & 48.00 & 50.67 & 69.00 & 73.00          & \textbf{75.33}  \\
                              & 151-175     & 88.44 & 67.17 & 64.15 & 73.87 & 81.07          & \textbf{83.75}  \\
                              & 176-200     & 87.90 & 83.70 & 82.02 & 85.21 & 87.23          & \textbf{88.40}  \\
                              & 1-200       & 76.75 & -     & -     & -     & \textbf{67.95} & 67.91           \\ 
\hline
\multirow{6}{*}{Cars196}      & 1-98        & 82.26 & -     & -     & -     & 73.76          & \textbf{75.54}  \\
                              & 99-123      & 97.13 & -     & -     & -     & 94.86          & \textbf{96.42}  \\
                              & 124-148     & 98.19 & -     & -     & -     & 96.38          & \textbf{96.86}  \\
                              & 149-173     & 97.38 & -     & -     & -     & 97.26          & \textbf{98.45}  \\
                              & 174-196     & 94.52 & -     & -     & -     & 95.54          & \textbf{96.56}  \\
                              & 1-196       & 85.15 & -     & -     & -     & 75.00          & \textbf{76.42}  \\ 
\hline
\multirow{6}{*}{DeepFashion2} & 1-10595     & 56.02 & -     & -     & -     & 55.16          & \textbf{56.04}  \\
                              & 10596-13256 & 67.30 & -     & -     & -     & 66.94          & \textbf{67.08}  \\
                              & 13257-15890 & 67.62 & -     & -     & -     & 67.56          & \textbf{68.33}  \\
                              & 15891-18558 & 68.42 & -     & -     & -     & 68.22          & \textbf{69.01}  \\
                              & 18559-21190 & 66.22 & -     & -     & -     & 67.41          & \textbf{67.76}  \\
                              & 1-21190     & 49.66 & -     & -     & -     & 49.78          & \textbf{50.58}  \\
\bottomrule
\end{tabular}}
\caption{Recall@1 (\%) on multi-task online learning. ``Joint'' (namely Joint Training) serves as the reference.}
\label{table:multitask-results}
\end{table}

As shown on Tab.~\ref{table:ablation-feature}, when the mutual learning is integrated with the feature estimation of the way in FECD~\cite{chen2021feature}, the performance of ``FECD(Mutual)'' is inferior to ours in most of the cases. Moreover, it can be observed that FECD(Mutual) performs better than FECD on the last three tasks, which reveals the effectiveness of mutual loss over the conventional teacher-student model.

\begin{table}[tb]
\centering
\scriptsize{
\begin{tabular}{cccccccc} 
\toprule
\multirow{2}{*}{Datasets}     & \multirow{2}{*}{Approaches} & \multicolumn{3}{c}{ResNet-50}                    & \multicolumn{3}{c}{Vision Transformer}            \\
                              &                          & Old            & New            & All            & Old            & New            & All             \\ 
\hline
\multirow{5}{*}{CUB-200}      & Initial                  & 79.53          & 33.53          & 50.42          & 83.81          & 65.00          & 71.12           \\
                              & Fine-tuning              & 51.10          & 79.62          & 57.81          & 77.25          & 80.59          & 75.46           \\
                              & Joint Training           & 79.07          & 78.49          & 76.50          & 82.97          & 81.97          & 80.57           \\
                              & FECD                     & 75.59          & 75.08          & 70.44          & \textbf{82.25} & 79.03          & 78.08           \\
                              & \textbf{Ours}            & \textbf{76.69} & \textbf{75.34} & \textbf{70.55} & 82.16          & \textbf{80.17} & \textbf{78.59}  \\ 
\hline
\multirow{5}{*}{Cars196}      & Initial                  & 85.21          & 41.81          & 57.49          & 80.63          & 52.87          & 60.68           \\
                              & Fine-tuning              & 62.49          & 94.44          & 71.16          & 57.58          & 87.81          & 65.66           \\
                              & Joint Training           & 84.99          & 93.86          & 88.24          & 77.56          & 86.01          & 78.45           \\
                              & FECD                     & \textbf{83.36} & 92.13          & 83.24          & \textbf{77.53} & 83.98          & 75.82           \\
                              & \textbf{Ours}            & 82.81          & \textbf{93.58} & \textbf{83.78} & 76.40          & \textbf{85.89} & \textbf{75.93}  \\ 
\hline
\multirow{5}{*}{DeepFashion2} & Initial                  & 53.06          & 50.87          & 45.65          & 58.40          & 58.36          & 52.51           \\
                              & Fine-tuning              & 52.70          & 53.30          & 46.57          & 58.37          & 58.14          & 52.34           \\
                              & Joint Training           & 54.03          & 53.99          & 47.66          & 58.40          & 58.46          & 52.49           \\
                              & FECD                     & \textbf{54.47} & 54.45          & 47.94          & 58.29          & 58.35          & 52.37           \\
                              & \textbf{Ours}            & 54.11          & \textbf{54.88} & \textbf{48.14} & \textbf{58.44} & \textbf{58.44} & \textbf{52.47}  \\
\bottomrule
\end{tabular}}
\caption{Ablation study with different backbones on one-task online learning. Evaluated by Recall@1 (\%).}
\label{table:ablation-backbone}
\end{table}

\begin{table}[tb]
\centering
\scriptsize{
\begin{tabular}{ccccc} 
\toprule
Tasks   & Mutual & FECD  & FECD(Mutual)          & Ours  \\ 
\hline
1-100   & 71.57  & 73.64 & \textbf{73.98}       & 73.90        \\
101-125 & 75.68  & \textbf{77.21} & 76.19       & \textbf{77.21}        \\
126-150 & 70.50  & 73.00 & 73.17                & \textbf{75.33}        \\
151-175 & 81.24  & 81.07 & 82.08                & \textbf{83.75}        \\
176-200 & 87.56  & 87.23 & 88.07                & \textbf{88.40}        \\
1-200   & 66.12  & \textbf{67.95} & 67.76       & 67.91        \\
\bottomrule
\end{tabular}}
\caption{Ablation study with different feature estimation approaches on CUB-200 on multi-task online learning. Evaluated by Recall@1 (\%).}
\label{table:ablation-feature}
\end{table}

\begin{table}[htb]
\centering
\scriptsize{
\begin{tabular}{ccccc} 
  \toprule
  Tasks   & Joint & FECD           & \textbf{Ours}   \\ 
  \hline
  1-100   & 82.97 & 80.51          & \textbf{81.40}  \\
  101-125 & 85.54 & 81.46          & \textbf{82.48}  \\
  126-150 & 77.33 & 78.17          & \textbf{78.50}  \\
  151-175 & 87.60 & 83.92          & \textbf{86.60}  \\
  176-200 & 90.42 & \textbf{88.91} & 88.40           \\
  1-200   & 80.57 & 76.62          & \textbf{77.83}  \\
  \bottomrule
\end{tabular}
}
\caption{Ablation study with Vision Transformer as the backbone on CUB-200 on multi-task online learning. Recall@1 (\%) is the evaluation metric.}
\label{table:ablation-vit-multitask}
\end{table}

\subsection{Ablation Study}
The performance of our approach is also studied when the BN-Inception backbone is replaced with ResNet-50 and Vision Transformer. The study is conducted on both one-task and multi-task scenarios. The one-task and multi-task online learning results are shown on Tab.~\ref{table:ablation-backbone} and Tab.~\ref{table:ablation-vit-multitask} respectively. Although the performance fluctuates considerably for all the approaches across different backbones, our approach outperforms the competing approach FECD in most of the cases. This confirms the stability of our approach across different backbones.

%We perform experiments on various ablations of different feature estimation methods to see the influence of each method. We compare the individual performance of the single mutual loss, FECD method, the combination of mutual loss and FECD, and the combination of mutual loss and PDFE (Ours) in multi-task online training. Table ~\ref{table:ablation-feature} shows the ablation study results on the CUB-200 dataset. The model performs best when our method is considered. It is observed that FECD also improves the performance on prior tasks and only applying mutual loss without feature estimation performs badly. In addition, we replace BN\hyp{}Inception with ResNet50~\cite{he2016deep} or Vision Transformer (ViT) ~\cite{dosovitskiy2020image} as backbone to see the influence of different backbones. Table ~\ref{table:ablation-backbone} shows the results on three datasets. In general, our method outperforms FECD in most cases.

%% file: conclude.tex
\section{Conclusion}
\label{sec:conc}
We have presented our solution for online deep metric learning both for one stage and multiple stage scenarios. Different from existing solutions, our approach is built upon a mutual learning structure, which turns out to make a good balance between the old and new learning tasks. Moreover, a novel virtual feature estimation approach is proposed. In combination with mutual learning, constantly superior performance is achieved in the multi-task learning scenario. In most cases, it even outperforms the joint training. The effectiveness of our approach is validated on different datasets and with different network backbones.
%In this work, we constrain feature correlations of new task under a mutual framework for online deep metric learning. In the multi-task scenario, instead of loading all prior models which increases the usage of memory, we propose an approach to estimate features of all prior tasks, which provides extra previous knowledge for the current task to alleviate the catastrophic forgetting. On three datasets, the efficacy of our approach is verified in comparison to other methods.